\documentclass[11pt]{article}

\usepackage[preprint]{acl}

\usepackage{times}
\usepackage{latexsym}

\usepackage[T1]{fontenc}
\usepackage[utf8]{inputenc}
\usepackage{microtype}
\usepackage{inconsolata}
\usepackage{graphicx}

\usepackage{booktabs}
\usepackage{amsfonts}
\usepackage{amsmath}
\usepackage{xcolor}
\usepackage{tcolorbox}
\usepackage{color,soul}
\usepackage{stfloats}

\newcommand{\agentname}{Eluna}

\title{\agentname: An Agentic LLM System for Automating Warehouse Operations with Reasoning and Task Execution}

\author{
\textbf{Ning Liu}\thanks{\ Equal contribution.} \quad \textbf{P Aditya Sreekar}\footnotemark[1] \quad \textbf{Kalle Kujanp\"a\"a}\footnotemark[1] \quad \textbf{Zhaoxuan Zhu}\footnotemark[1] \quad \textbf{Kaiwen Liu} \\
\textbf{Chuanneng Sun} \quad \textbf{Jorge Marchena Menendez} \quad \textbf{Matthew Bales} \quad \textbf{Tianyu Yang} \\
\textbf{Shahnawaz Alam} \quad \textbf{Rose Yu} \quad \textbf{Baoyuan Liu} \quad \textbf{Kristina Klinkner} \quad \textbf{Shervin Malmasi}
\\\\
Amazon.com, Inc. \quad Fulfillment Technologies and Robotics
\\
\texttt{\{ningliun,malmasi\}@amazon.com}
}

\begin{document}
\maketitle
\begin{abstract}
Warehouse operations are governed by Standard Operating Procedures (SOPs) that encode complex, multi-system decision logic, which must be executed reliably under strict time constraints, yet LLM agents lack mechanisms to enforce procedural compliance and degrade under the context overload full SOP specifications introduce. We present \agentname{}, a production-deployed agentic system for reliable SOP execution. \agentname{} is a graph-guided, multi-agent framework that encodes SOPs as directed acyclic graphs with progressive disclosure and delegates independent tasks to parallel sub-agents, each with persistent code execution and live data access. To meet production latency and accuracy needs, we use asymmetric episodic distillation where a strong teacher is improved through episodic error memories, then a smaller student is fine-tuned on the corrected trajectories with memory stripped, internalizing corrections without inference-time overhead. On a 13-task benchmark and two production applications, our fine-tuned models match or exceed their teacher, beat all larger off-the-shelf baselines, and reach 94\% expert agreement on the ticket processing application.

\end{abstract}

\section{Introduction}

Modern warehouse operations require continuous monitoring, multi-step diagnosis, and timely intervention across dozens of systems, ranging from robotic conveyance operators tracing threshold breaches to root causes within minutes, to inventory ticket processing that queries state, validates constraints, and submits physical item picks across many systems per ticket.  Robotic conveyance diagnosis and ticket processing are two among many such workflows, each governed by Standard Operating Procedures (SOPs): directed decision pathways with prescribed steps, dependencies, quantitative thresholds, and constraints that must be followed faithfully. Today much of their execution is manual, slow, and error-prone, and delays in diagnosis directly degrade throughput, making automated SOP execution a pressing business need.

Large language models reason well~\cite{wei2022chain,kojima2022large}, but agentic frameworks such as ReAct~\cite{yao2023react} and Reflexion~\cite{shinn2024reflexion} rely on loosely structured prompts with no mechanism to strongly guide SOP compliance, so a capable model can still deviate from prescribed decision paths. Recent benchmarks confirm this gap on workflow-guided tasks~\cite{xiao2024flowbench,wang2025sop}, industrial SOP-following~\cite{nandi2025sop}, and operational diagnostics where even SOP-enhanced multi-agent systems plateau in performance ~\cite{pei2025flow}. A core bottleneck is context overload: with the full SOP in view, performance degrades as complex workflows overwhelm the model's ability to select relevant actions~\cite{xiao2024flowbench,pei2025flow}.
This plateau holds regardless of model scale, and prompting alone is insufficient without domain-specific training \cite{nandi2025sop}.

We present a graph-guided agent framework that addresses these challenges through a unified execution model. SOPs are encoded as directed acyclic graphs, and progressive disclosure surfaces only the reachable subgraph and node-level specifications on demand. A main agent orchestrates traversal and delegates independent node evaluations to parallel sub-agents in isolated contexts, each with a persistent code interpreter (the CodeAct paradigm of \citet{wang2024executable}) and access to real-time data via the Model Context Protocol~\cite{anthropic2024mcp}, providing context isolation and parallelism while guiding the agent toward procedural compliance. Together, progressive disclosure and sub-agent isolation address context overload. Each operational use case is packaged as a self-contained \emph{Skill}\footnote{\url{https://agentskills.io/}: a bundle comprising the decision graph, use-case-specific tools, node specifications, and execution instructions.} loaded on demand, so new workflows need no bespoke systems. Since prompting alone is insufficient~\cite{nandi2025sop}, we pair the framework with a trajectory-centric training pipeline where episodic learning corrects a strong teacher's trajectories without weight updates, and a smaller, low-latency student is fine-tuned without episodic memory (asymmetric episodic distillation), internalizing the episodic corrections in its weights rather than depending on them at inference. We make three contributions:
\begin{itemize}
    \item A \textbf{graph-guided skill-based agent framework} that encodes SOPs as directed acyclic graphs with progressive disclosure, and uses hierarchical multi-agent execution with parallel sub-agent delegation for scalable, procedurally compliant reasoning across multiple operational use cases.
    \item A \textbf{trajectory-centric training pipeline with asymmetric episodic distillation}: episodic learning iteratively improves teacher trajectory quality, and the student is fine-tuned on per-turn decomposed trajectories filtered by rejection sampling, with episodic memory stripped, internalizing the episodic corrections in its weights and eliminating inference-time memory dependence.
    \item \textbf{Real-world performance} on a 13-task operational reasoning benchmark and two production warehouse applications (robotic conveyance and ticket processing), demonstrating that a fine-tuned 32B model outperforms all baselines including its teacher and larger off-the-shelf models.  A fine-tuned 355B model achieves a 94\% human-match on the ticket processing application.
\end{itemize}

\section{Related Work}

\paragraph{Tool-Augmented Agents and Multi-Step Reasoning.}
LLM agents have been equipped with code execution~\cite{gao2023pal,wang2024executable}, APIs~\cite{patil2023gorilla,qin2024toolllm}, and standardized tool protocols~\cite{anthropic2024mcp}. Multi-step reasoning methods range from chain-of-thought~\cite{wei2022chain} and tree-of-thoughts~\cite{yao2024tree} to interleaved reasoning and action~\cite{yao2023react} and hierarchical decomposition~\cite{khot2023decomposed,besta2024graph}. These approaches improve general reasoning but rely on loosely structured prompts without mechanisms to enforce deterministic procedural constraints.

\paragraph{Structured Reasoning and Operational AI.}
Multi-agent frameworks such as AutoGen~\cite{wu2023autogen} and MetaGPT~\cite{hong2024metagpt} enable task decomposition across specialized agents. SOPStruct~\cite{garg2025sopstruct} converts unstructured SOP text into DAG representations, but is an upstream \emph{structuring} system: its output is a structured graph, and its evaluation measures structuring quality (completeness and soundness of the DAG). It has no execution engine, tool use, or assessment of task completion, so it occupies a different pipeline stage than \agentname{}, which addresses the downstream problem of faithfully \emph{executing} a structured SOP graph in production; a direct empirical comparison is therefore not meaningful, and the two are complementary. Agent-S~\cite{kulkarni2025agents} navigates SOPs via prompt-driven state machines but lacks graph-guided progressive disclosure, code execution, and post-training; we compare empirically against this paradigm in Section~\ref{sec:ablation}. Our work integrates SOP graphs into both inference and training.

\paragraph{Trajectory-Centric Training.}
Reflexion~\cite{shinn2024reflexion} and Self-Refine~\cite{madaan2024selfrefine} use iterative feedback to improve trajectories at inference time. FireAct~\cite{chen2023fireact} and AgentTuning~\cite{zeng2024agenttuning} show that fine-tuning on agent trajectories improves generalization. Our pipeline differs in two ways: episodic learning improves the \emph{teacher} trajectories used for distillation rather than the deployed model, and training is aligned with graph-structured procedural logic rather than free-form tasks.

\section{A Graph-Guided Operational Agent}
\label{sec:framework}

\begin{figure*}[!t]\centering
\includegraphics[width=1.0\linewidth]{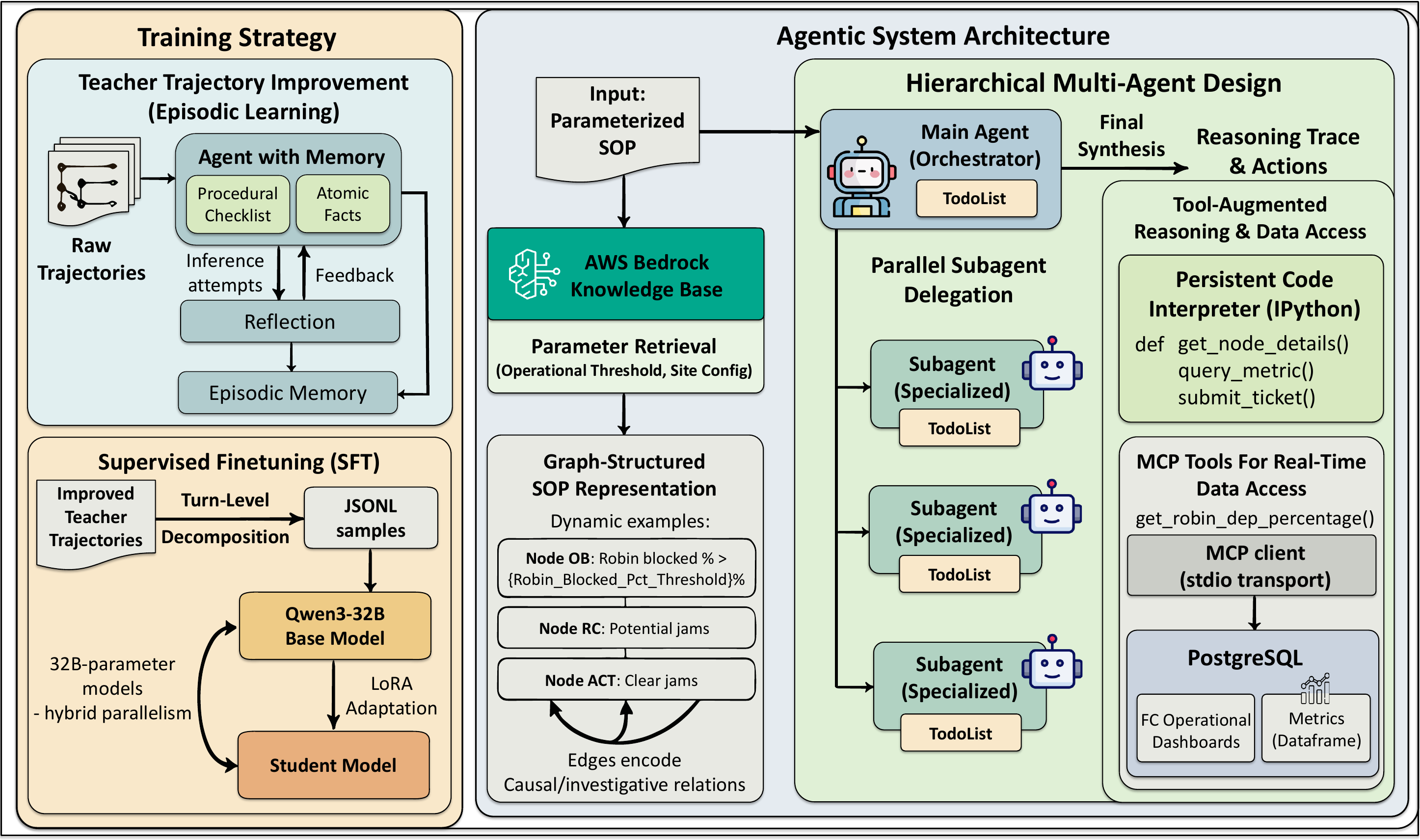}
 \caption{Overview of the \agentname{} framework. \textbf{Left:} Training strategy combining episodic learning for teacher trajectory improvement with turn-level decomposition for supervised fine-tuning via LoRA. \textbf{Right:} system architecture featuring graph-structured SOP representation, hierarchical multi-agent design with parallel sub-agent delegation, and tool-augmented reasoning through persistent code execution and MCP-based data access.}
 \label{fig:architecture}
\end{figure*}

\agentname{} is a hierarchical multi-agent framework that addresses the two challenges above (context overload and use-case generality) through graph-structured progressive disclosure, parallel sub-agent delegation, and skill packaging. Figure~\ref{fig:architecture} provides an overview.

\subsection{Graph-Structured SOP Representation}
\label{subsec:graph}

Each SOP is a directed acyclic graph $G = (V, E)$ with five node types: \textit{observations} (detectable conditions), \textit{root causes} (diagnosed causes), \textit{calculations} (metric analysis), \textit{constraints} (preconditions), and \textit{actions} (remediation). Each node has a parameterized specification defining its evaluation procedure, which produces a structured outcome: a boolean, a classification (e.g., a category that routes downstream traversal), or a computed value. Edges encode directed dependencies: a single-node edge activates the child on the boolean outcome of one parent (holding, or not holding for fallback paths), while a multi-node edge requires several parents to hold simultaneously. A node may have multiple outgoing edges.

\paragraph{Progressive Disclosure.} Presenting the full graph degrades traversal accuracy~\cite{liu2024lost}, especially with parallel paths, where the agent conflates branches, omits nodes, or explores irrelevant subtrees. We disclose the graph at two levels. First, a retrieval tool runs BFS from the triggered node and returns only the reachable subgraph (node types, edge relationships, brief descriptions). Second, each node's full procedural specification (evaluation logic, required tool calls, thresholds) is fetched only when the agent begins evaluating it, bounding working context to one node at a time.

\subsection{Hierarchical Multi-Agent Architecture}
\label{subsec:delegation}

\agentname{} uses a two-level hierarchy: a main agent orchestrates traversal while sub-agents evaluate individual nodes in parallel. The main agent holds the retrieved subgraph, selects the next nodes to evaluate from edge dependencies, and issues delegation calls; each sub-agent receives a fresh context with only the assigned node's procedural specification, role instructions, and tools. Action nodes remain with the main agent because they require graph-level state for sequencing and constraint checking.

Multiple delegation calls issued in one turn execute in parallel, reducing wall-clock time to roughly the slowest single evaluation. Only each sub-agent's structured conclusion is returned, so the main agent's context grows by one summary per node rather than the full investigation history; sub-agent sessions persist so the main agent can query one for detail without re-running it, and sub-agents cannot spawn further sub-agents. This isolates tool outputs and intermediate computation within sub-agent sessions, addressing context overload by construction while the main agent holds only graph structure and node conclusions. In a parallel effort, we reduce the per-node sub-agent overhead by compiling repeated SOP steps into reusable tools at build time~\cite{kujanpaa2026toolmaking}.

\subsection{Skill Packaging}
\label{subsec:skills}

Each warehouse workflow needs its own decision logic, prompts, data sources, and actions. A custom agent per use case duplicates the execution engine and training pipeline, while a single agent loaded with all use cases suffers context overload and tool-space conflicts. The workflow is a more natural packaging unit: we bundle per-workflow artifacts into a self-contained \emph{skill} loaded at runtime, following the standardized skill format models are increasingly trained to use.
The framework stays use-case-agnostic so the same agent and execution logic serve any loaded skill, and adding a workflow means authoring a skill, not modifying the agent.

A skill comprises five components: (1)~the decision graph encoding the SOP as a DAG, (2)~use-case-specific tools accessed via the Model Context Protocol~\cite{anthropic2024mcp} for querying data and executing actions, (3)~node detail specifications containing per-node evaluation logic, (4)~CodeAct functions~\cite{wang2024executable}, and (5)~execution instructions directing traversal strategy (e.g., when to delegate, how to handle constraints); reference documents for knowledge retrieval~\cite{schick2024toolformer} are bundled as readable files. When a trigger fires, a skill-loading tool registers the skill's tools under a namespace prefix, injects its CodeAct functions into the persistent interpreter as importable modules, and appends its execution instructions, after which the agent has exactly the tools and instructions for the current use case.

\paragraph{Automated SOP Authoring.} Skills do not require manual graph authoring. A separate LLM-based authoring agent, developed in parallel, extracts operational knowledge from existing documents and subject-matter-expert (SME) interviews, converts it into the formal graph representation (identifying node types, edge dependencies, and evaluation logic), and compiles the result into an executable skill package. A validation loop closes the process: \agentname{} generates reasoning traces against the compiled skill, disagreements with expected outcomes are surfaced to the SME for feedback that drives targeted graph refinements, and the loop iterates until agreement metrics meet a predefined threshold, at which point the skill is deployed. This authoring pipeline is orthogonal to the execution framework and training strategy that are the focus of this paper, so we do not detail it further, but it makes the approach scalable to new workflows.

\subsection{Tool Ecosystem}
\label{subsec:tools}
The agent operates over five tools: a persistent code interpreter
for programmatic data manipulation and metric computation; a task-tracking list that doubles as a steering mechanism to keep traversal on the intended path; MCP-based data and action access~\cite{anthropic2024mcp} to operational dashboards; agentic RAG~\cite{schick2024toolformer} for on-demand knowledge retrieval; and an operative memory that persists across invocations, accumulating knowledge and enabling deduplication of redundant investigations. Appendix~\ref{app:tools} details each one.

\begin{table*}[t]
\centering
\setlength{\tabcolsep}{4pt}
\resizebox{0.99\textwidth}{!}{
\begin{tabular}{clrrrrr}
\toprule
 & \textbf{Task} & \textbf{GLM-4.7} & \textbf{Qwen3-235B} & \textbf{Qwen3-32B} & \textbf{\agentname{}-Q} & \textbf{\agentname{}-G} \\
\midrule
1  & Metric Reading        & +6.6  & -1.1  & -57.1 & +5.2  & \textbf{+6.6}  \\
2  & State Classification        & \textbf{+4.3}  & -22.3 & -34.0 & +2.4  & +2.1  \\
3  & Operational Language Understanding      & \textbf{+4.4}  & -12.1 & -7.7  & +4.4  & +1.1  \\
4  & Metric Trend Detection       & +11.9 & +9.5  & -36.9 & +10.7 & \textbf{+14.3} \\
5  & Causal Chain Analysis        & \textbf{+25.0} & -22.4 & -38.2 & +13.2 & +23.7 \\
6  & Statistical Pattern Analysis       & -5.0  & +1.2  & -80.0 & \textbf{+9.1}  & 0.0   \\
7  & Knowledge Base Lookup           & +3.4  & -20.7 & -8.0  & \textbf{+4.6}  & +3.4  \\
8  & Query Understanding      & +11.1 & -13.9 & -31.9 & \textbf{+19.0} & +12.5 \\
9  & Cross-System Knowledge Synthesis   & +6.3  & \textbf{+15.9} & -12.7 & +7.9  & 0.0   \\
10 & Complex Query Decomposition        & -2.2  & -21.3 & -37.1 & \textbf{+1.1}  & -1.1  \\
11 & Temporal Comparison      & \textbf{+14.3} & -8.6  & -77.1 & +2.9  & +14.3 \\
12 & Multi-Factor Context Awareness        & +2.2  & -5.6  & -31.5 & -3.4  & \textbf{+7.9}  \\
13 & Knowledge Base Context Filtering    & +3.3  & -20.9 & -45.1 & \textbf{+4.7}  & +3.3  \\
\midrule
\multicolumn{2}{l}{\textbf{AVG}} & +6.1 & -10.0 & -38.0 & +6.0 & \textbf{+6.5} \\
\bottomrule
\end{tabular}
}
\caption{Operational benchmark: \% improvement over the GLM-4.5-Air baseline across 13 tasks.}
\label{tab:benchmark_main}
\end{table*}

\section{Training Strategy}

The graph-guided framework (Section~\ref{sec:framework}) is model-agnostic, but even the largest available models fail to reach reliable accuracy on graph-structured procedural reasoning~\cite{nandi2025sop}, motivating domain-specific training. Our trajectory-centric pipeline has three stages. First, a strong teacher generates trajectories within the full framework, and episodic learning (EL)~\cite{shinn2024reflexion}. EL is a text memory of past errors injected into the prompt, which iteratively corrects the teacher's mistakes without weight updates. Second, rejection sampling~\cite{yuan2023scaling} with LLM-judge evaluation~\cite{zheng2024judging} retains only correct trajectories. Third, a smaller student that meets the production latency SLA is fine-tuned on these trajectories \emph{with the EL memory stripped}. We call this \emph{asymmetric} episodic distillation because EL is applied only on the teacher side and the student never sees the memory during training or inference, so it must internalize the corrections in its weights, and no runtime memory store is needed. Appendix~\ref{app:training_pipeline} details each stage.

\section{Experiments}

\subsection{Experimental Setup}
\label{subsec:setup}

The agent framework is implemented using the Strands Agents SDK~\cite{strands2025}, following the design in Section~\ref{sec:framework}. We fine-tune Qwen3-32B~\cite{yang2025qwen3} and GLM-4.5-Air~\cite{zeng2025glm} with LoRA~\cite{hu2022lora} on 8,525 training samples (expanded from trajectories via turn-level decomposition, Appendix~\ref{app:training_pipeline}). Training and inference details are in Appendix~\ref{app:training_details}.
\paragraph{Datasets}
We evaluate on three tasks: an \textbf{operational reasoning benchmark} of 13 task types testing capabilities required for warehouse operational reasoning (1,400 evaluation samples; Table~\ref{tab:benchmark_main}), a \textbf{robotic conveyance application} covering end-to-end SOP execution over 1,500 real scenarios, and a \textbf{ticket processing application} executing a 46-node inventory consolidation SOP on 410 real tickets evaluated against expert annotations. For the first two, ground truth is generated from real operational data by sampling timestamps, checking root-node conditions, and running a deterministic SOP implementation where they hold, with temporal partitioning separating training from evaluation. Ticket processing has no deterministic ground truth, so correctness is estimated from expert review of sampled agent traces.

\paragraph{Evaluation}
Claude Sonnet 4 is used as the automated judge of answer correctness for the first two tasks. On the benchmark, the judge scores model outputs against reference answers. On the robotic conveyance application, node classifications are extracted from agent trajectories and compared against ground-truth labels, reporting binary correctness (per-node accuracy) and exact match (all nodes of a type correct within a scenario). Because ground truth is deterministic, judging reduces to structured matching rather than open-ended assessment; the LLM judge was validated against human judgements and achieves full agreement, confirming its reliability for this structured evaluation setting. For ticket processing, correctness is estimated from expert review of agent traces (Appendix~\ref{app:ciss}). Results are reported as improvement over the GLM-4.5-Air off-the-shelf (OTS) baseline.

\paragraph{Models}
All models are evaluated within the same framework, with identical tools and skills and with episodic memory stripped at inference (Section~\ref{sec:framework}). The \textbf{baseline} is OTS GLM-4.5-Air, a 106B parameter model that meets the production latency SLA, hence the reference point for what is achievable under operational constraints. GLM-4.7 (355B), the \textbf{teacher} that generated training trajectories, is too large to meet that SLA, so it serves as a trajectory generator and upper reference rather than a deployable baseline. GLM-4.7 was selected as teacher because it achieved the highest off-the-shelf accuracy on our benchmark among the open-weight models we evaluated, and its reasoning patterns align with our target student (GLM-4.5-Air), facilitating knowledge transfer; distillation from closed commercial models is not permitted under their terms of service, constraining teacher selection to open weights. The distillation framework itself is model-agnostic, and any sufficiently capable open-weight teacher can be substituted. We additionally compare OTS Qwen3-235B-A22B-Thinking and OTS Qwen3-32B, against our two fine-tuned models \agentname{}-Q (Qwen3-32B) and \agentname{}-G (GLM-4.5-Air). For the ticket task, which runs on a longer interval, we additionally fine-tune the teacher itself into \agentname{}-G-XL, whose latency meets that SLA.

\subsection{Results}
\label{sec:benchmark_results}

\paragraph{Operational Benchmark.}

Table~\ref{tab:benchmark_main} reports per-task results. \agentname{}-G improves 6.5\% and \agentname{}-Q improves 6.0\% over the baseline on average. \agentname{}-Q, trained on GLM-4.7 trajectories, matches the teacher (+6.0\% vs.\ +6.1\%) despite being a different and smaller architecture, showing that our pipeline transfers operational reasoning across model families, while OTS Qwen3-32B averages 38.0\% below the baseline, showing the gap that domain-specific training closes. The largest gains concentrate on complex multi-step tasks like Causal Chain Analysis (Task~5), Query Understanding (Task~8), and Query Decomposition (Task~10). These require chaining tool calls and synthesizing across data sources, indicating that training instills operational reasoning rather than single-step retrieval.

\begin{table}[tb]
\centering
\footnotesize
\resizebox{\columnwidth}{!}{%
\begin{tabular}{lrrrrrr}
\toprule
& \multicolumn{3}{c}{\textbf{Binary Correctness}} & \multicolumn{3}{c}{\textbf{Exact Match}} \\
\cmidrule(lr){2-4} \cmidrule(lr){5-7}
\textbf{Model} & OB & RC & ACT & OB & RC & ACT \\
\midrule
GLM-4.7 (teacher)   & +50.2 & +73.7 & +8.5 & +93.7 & +84.0 & +164.8 \\
\agentname{}-Q    & +48.9 & +73.9 & +8.5 & +93.7 & \textbf{+84.4} & \textbf{+165.3} \\
\agentname{}-G  & +49.5 & \textbf{+73.9} & +8.5 & \textbf{+93.9} & \textbf{+84.4} & +164.8 \\
\bottomrule
\end{tabular}%
}
\caption{Robotic conveyance application: \% improvement over GLM-4.5-Air baseline, averaged over 4 runs.}
\label{tab:application_results}
\end{table}

\paragraph{Robotic Conveyance Application.}
End-to-end SOP execution is far harder than isolated benchmark tasks: even with full framework access, the OTS baseline compounds per-node errors across the traversal and degrades sharply. We report binary correctness and exact match (Section~\ref{subsec:setup}) per node type (observations, root causes, and actions) to reflect the distinct sub-tasks of detection, diagnosis, and remediation. Fine-tuning on complete trajectories closes the gap (Table~\ref{tab:application_results}): all trained models improve observation binary correctness by over +49\%, root-cause binary correctness by over +73\%, and exact match by +84\% on root causes and +165\% on actions, with our 32B model matching the GLM-4.7 teacher across all metrics, which shows that trajectory distillation transfers full SOP traversal behavior, not just isolated skills.

Median end-to-end execution latency (wall-clock from query to final action) is also reduced by 54.3\% (\agentname{}-Q) and 54.8\% (\agentname{}-G) relative to GLM-4.7, meeting the latency constraints for real-time diagnostic support.

\subsection{Ablation Studies}
\label{sec:ablation}

\paragraph{Architectural Ablation.}
Table~\ref{tab:ablation_arch} isolates the contribution of the two architectural components of Section~\ref{sec:framework}, namely progressive disclosure and multi-agent delegation, by running GLM-4.7 on the robotic conveyance application under all four configurations. The single-agent, full-SOP configuration doubles as a paradigm comparison: it simulates Agent-S~\cite{kulkarni2025agents}, where a single LLM receives the entire SOP as a prompt-driven state machine without progressive disclosure or parallel delegation.

\pagebreak

\begin{table}[tb]
\centering
\footnotesize
\resizebox{\columnwidth}{!}{%
\begin{tabular}{lrrr}
\toprule
\textbf{Configuration} & \textbf{OB} & \textbf{RC} & \textbf{ACT} \\
\midrule
Single-agent, full SOP (Agent-S style) & +76.1 & +73.3 & +129.5 \\
Multi-agent, full SOP                  & +81.0 & +77.6 & +137.1 \\
Single-agent, progressive disclosure   & +91.6 & +77.7 & +158.0 \\
Multi-agent, prog.\ disclosure (\agentname{}) & \textbf{+93.9} & \textbf{+84.4} & \textbf{+164.8} \\
\bottomrule
\end{tabular}%
}
\caption{Architectural ablation on the robotic conveyance application (All rows use GLM-4.7): exact-match \% improvement over the GLM-4.5-Air baseline, as in Table~\ref{tab:application_results}.}
\label{tab:ablation_arch}
\end{table}

Both components contribute meaningfully. Progressive disclosure provides the larger individual gain (rows 1 vs.\ 3: +76.1 to +91.6 on observations, +129.5 to +158.0 on actions), consistent with the context-overload hypothesis of Section~\ref{subsec:graph}. Sub-agent delegation adds complementary improvements, particularly on actions (+158.0 to +164.8), where isolated node evaluation benefits graph-level state management. The full \agentname{} architecture achieves the best results across all metrics and outperforms the Agent-S-style configuration by +17.8 on observations, +11.1 on root causes, and +35.3 on actions, confirming that the gains come from how the SOP graph is integrated into inference rather than from model strength alone.

\paragraph{Episodic Learning Ablation.}
Table~\ref{tab:ablation_el} isolates the contribution of EL by varying where the EL memory is applied. We use Tasks~5 and~10 from the benchmark as a deliberate proof of concept before full-scale training: they are the hardest tasks for GLM-4.7 (Table~\ref{tab:benchmark_main}), maximizing the room for episodic correction to demonstrate measurable gains; the full pipeline (Tables~\ref{tab:benchmark_main} and~\ref{tab:application_results}) then applies the validated design across all 13 tasks and both production applications. Results are reported as percentage change relative to the no-EL variant.

\begin{table}[tb]
\centering
\footnotesize
\resizebox{\columnwidth}{!}{%
\begin{tabular}{lrr}
\toprule
\textbf{Setting} & \textbf{Task 5 (\%)} & \textbf{Task 10 (\%)} \\
\midrule
No EL anywhere (reference)               & 0.0   & 0.0   \\
EL in teacher + student (train \& eval)  & +16.6 & +25.3 \\
EL in teacher + student (train only)     & -52.7 & -40.0 \\
EL in teacher only (\agentname{})                & \textbf{+16.2} & \textbf{+28.6} \\
\bottomrule
\end{tabular}%
}
\caption{Episodic learning (EL) ablation: \% change relative to the no-EL variant.}
\label{tab:ablation_el}
\end{table}

Applying EL only to the teacher (our approach) yields +16.2\% on Task~5 and +28.6\% on Task~10, matching the variant that also feeds EL to the student at inference (Task~5: +16.6\%, Task~10: +25.3\%) but without its runtime memory dependence. Training the student with EL yet removing it at inference degrades catastrophically (Task~5: $-$52.7\%, Task~10: $-$40.0\%), confirming that our asymmetric design must distill episodic knowledge into parameters rather than expose it as a brittle inference-time dependency. Appendix~\ref{app:ablation} details the four variants.

\section{Deployment and Business Impact}

Our framework is deployed in production across multiple operational applications, served on Amazon Bedrock AgentCore~\cite{aws_bedrock_agentcore_2026}. The two applications below apply it in contrasting regimes: robotic conveyance is latency-critical metric analysis, while ticket processing is latency-insensitive but needs complex reasoning.

\subsection{Robotic Conveyance}

This metric-analysis task queries live dashboards to compute metrics, and compares them against thresholds over a 63-node SOP graph whose high branching factor drives parallel sub-agent delegation. A degraded station can back up the main conveyor and erode facility-wide throughput in minutes, so diagnosis must be fast and correct.

In production deployment at a warehouse, the agent has processed roughly 8{,}000 triggers, producing 3{,}435 process alerts and 1{,}078 maintenance tickets. SOP adherence is near-perfect: SME review of agent trajectories finds the execution correct. The remaining errors trace to gaps in the SOP specification and upstream data quality rather than to agent reasoning. Prior to deployment, operators manually interpreted dashboards, navigated SOPs, and coordinated corrective actions. Early deployment feedback indicates a $300\%$ speedup in resolution time while reducing cognitive load and improving SOP adherence consistency.

\subsection{Ticket Processing}
\label{subsec:ticket}

\begin{table}[tb]
\centering
\resizebox{0.9\columnwidth}{!}{
\begin{tabular}{lr}
\toprule
\textbf{Metric} & \textbf{Value} \\
\midrule
Overall estimated correctness & 94.4\% \\
Consolidation decision precision & 96.7\% \\
Consolidation decision recall & 99.2\% \\
Consolidation decision F1 & 98.0\% \\
\midrule
Terminal distribution & \\
\quad Proceed to consolidation & 84.1\% \\
\quad Rejected (no eligible inventory) & 13.7\% \\
\quad Manual handling (outside SOP scope) & 2.2\% \\
\bottomrule
\end{tabular}
}
\caption{Ticket processing: \agentname{}-G-XL (open-weight 355B fine-tune) correctness (Appendix~\ref{app:ciss}).}
\label{tab:ciss_results}
\end{table}

This is a language-reasoning task. The agent interprets free-text ticket requests (expiration, damage, recalls, audits) and performs multi-step reasoning over structured inventory records (validating constraints, resolving eligibility, and deciding consolidation) rather than merely parsing the request. Each request type routes differently, and eligibility depends jointly on item condition and storage-location type. The SOP graph contains 46 nodes and is predominantly sequential, with mutually exclusive conditional paths. The network processes tens of thousands of these tickets daily. Approximately 80\% are handled through an existing scanning workflow. The remaining 20\% require manual desk processing at $\approx$6 minutes per ticket. We deploy \agentname{}-G-XL, an open-weight 355B fine-tune from the same pipeline. The agent's median generation duration is 187s (p90: 257s, mean: 210s), below the manual rate.

The agent automates the digital triage preceding physical inspection: interpreting the request, querying inventory, validating constraints, and submitting the pick, leaving the associate only the inspection. Precision (Table~\ref{tab:ciss_results}) is the operationally critical metric, as a false consolidation submits a pick that cannot be processed, wasting labor and blocking the queue. Only a small fraction of tickets need manual handling, freeing capacity for substantially larger volumes.

\section{Conclusion}

We presented \agentname{}, a production-deployed agentic system for reliable SOP execution that combines graph-guided progressive disclosure, hierarchical multi-agent delegation, and skill packaging with a trajectory-centric training pipeline built on asymmetric episodic distillation. Deploying \agentname{} in production surfaced two lessons that generalize beyond our setting. First, within the graph-guided framework, distillation beats scale for procedural compliance: a fine-tuned 32B model matches a far larger teacher and surpasses every OTS model tested, which plateau regardless of size. Second, the operational constraint of low latency under a production SLA is what forced the asymmetric distillation design, since exposing episodic memory at inference proved a brittle dependency rather than just an accuracy gain. The residual errors we observe trace to SOP-specification gaps and upstream data quality rather than agent reasoning, pointing to the SOP authoring and data-integrity pipeline as the next bottleneck for reliable operational automation.

\section*{Limitations}

Our system is currently deployed on two operational domains; while the skill-based architecture is designed for generality, transfer to substantially different operational environments remains to be validated empirically. The episodic learning pipeline assumes access to a capable teacher model and sufficient compute for multiple trajectory generation rounds, which may limit adoption in resource-constrained settings. Finally, both episodic learning and fine-tuning rely on ground-truth labels, which are expensive and time-consuming to obtain at scale; reducing this dependence is important future work.

\bibliography{refs}

@inproceedings{wei2022chain,
  title={Chain-of-Thought Prompting Elicits Reasoning in Large Language Models},
  author={Wei, Jason and Wang, Xuezhi and Schuurmans, Dale and Bosma, Maarten and Ichter, Brian and Xia, Fei and Chi, Ed and Le, Quoc and Zhou, Denny},
  booktitle={Advances in Neural Information Processing Systems},
  volume={35},
  year={2022}
}

@article{zeng2025glm,
  title={Glm-4.5: Agentic, reasoning, and coding (arc) foundation models},
  author={Zeng, Aohan and Lv, Xin and Zheng, Qinkai and Hou, Zhenyu and Chen, Bin and Xie, Chengxing and Wang, Cunxiang and Yin, Da and Zeng, Hao and Zhang, Jiajie and others},
  journal={arXiv preprint arXiv:2508.06471},
  year={2025}
}

@inproceedings{kojima2022large,
  title={Large Language Models are Zero-Shot Reasoners},
  author={Kojima, Takeshi and Gu, Shixiang Shane and Reid, Machel and Matsuo, Yutaka and Iwasawa, Yusuke},
  booktitle={Advances in Neural Information Processing Systems},
  volume={35},
  year={2022}
}

@article{garg2025sopstruct,
  title={Generating Structured Plan Representation of Procedures with {LLMs}},
  author={Garg, Deepeka and Zeng, Sihan and Ganesh, Sumitra and Ardon, Leo},
  journal={arXiv preprint arXiv:2504.00029},
  year={2025}
}

@article{kulkarni2025agents,
  title={Agent-{S}: {LLM} Agentic Workflow to Automate Standard Operating Procedures},
  author={Kulkarni, Mandar},
  journal={arXiv preprint arXiv:2503.15520},
  year={2025}
}

@article{yang2025qwen3,
  title={Qwen3 Technical Report},
  author={Yang, An and Yang, Baosong and Zhang, Beichen and others},
  journal={arXiv preprint arXiv:2505.09388},
  year={2025}
}

@inproceedings{yao2024tree,
  title={Tree of Thoughts: Deliberate Problem Solving with Large Language Models},
  author={Yao, Shunyu and Yu, Dian and Zhao, Jeffrey and Shafran, Izhak and Griffiths, Thomas L and Cao, Yuan and Narasimhan, Karthik},
  booktitle={Advances in Neural Information Processing Systems},
  volume={36},
  year={2024}
}

@misc{anthropic2024mcp,
  title={Model Context Protocol},
  author={{Anthropic}},
  year={2024},
  howpublished={\url{https://modelcontextprotocol.io}},
}

@inproceedings{besta2024graph,
  title={Graph of Thoughts: Solving Elaborate Problems with Large Language Models},
  author={Besta, Maciej and Blach, Nils and Kubicek, Ales and Gerstenberger, Robert and Podstawski, Michal and Gianinazzi, Lukas and Gajda, Joanna and Lehmann, Tomasz and Niewiadomski, Hubert and Nyczyk, Piotr and Hoefler, Torsten},
  booktitle={AAAI Conference on Artificial Intelligence},
  year={2024}
}

@inproceedings{khot2023decomposed,
  title={Decomposed Prompting: A Modular Approach for Solving Complex Tasks},
  author={Khot, Tushar and Trivedi, Harsh and Finlayson, Matthew and Fu, Yao and Richardson, Kyle and Clark, Peter and Sabharwal, Ashish},
  booktitle={International Conference on Learning Representations},
  year={2023}
}

@inproceedings{yao2023react,
  title={{ReAct}: Synergizing Reasoning and Acting in Language Models},
  author={Yao, Shunyu and Zhao, Jeffrey and Yu, Dian and Du, Nan and Shafran, Izhak and Narasimhan, Karthik and Cao, Yuan},
  booktitle={International Conference on Learning Representations},
  year={2023}
}

@inproceedings{schick2024toolformer,
  title={Toolformer: Language Models Can Teach Themselves to Use Tools},
  author={Schick, Timo and Dwivedi-Yu, Jane and Dess{\`\i}, Roberto and Raileanu, Roberta and Lomeli, Maria and Hambro, Eric and Zettlemoyer, Luke and Cancedda, Nicola and Scialom, Thomas},
  booktitle={Advances in Neural Information Processing Systems},
  volume={36},
  year={2024}
}

@article{patil2023gorilla,
  title={Gorilla: Large Language Model Connected with Massive APIs},
  author={Patil, Shishir G and Zhang, Tianjun and Wang, Xin and Gonzalez, Joseph E},
  journal={arXiv preprint arXiv:2305.15334},
  year={2023}
}

@inproceedings{qin2024toolllm,
  title={{ToolLLM}: Facilitating Large Language Models to Master 16000+ Real-world {APIs}},
  author={Qin, Yujia and Liang, Shihao and Ye, Yining and Zhu, Kunlun and Yan, Lan and Lu, Yaxi and Lin, Yankai and Cong, Xin and Tang, Xiangru and Qian, Bill and Zhao, Sihan and Hong, Lauren and Tian, Runchu and Xie, Ruobing and Zhou, Jie and Gerstein, Mark and Li, Dahai and Liu, Zhiyuan and Sun, Maosong},
  booktitle={International Conference on Learning Representations},
  year={2024}
}

@inproceedings{wang2024executable,
  title={Executable Code Actions Elicit Better {LLM} Agents},
  author={Wang, Xingyao and Chen, Yangyi and Yuan, Lifan and Zhang, Yizhe and Li, Yunzhu and Peng, Hao and Ji, Heng},
  booktitle={International Conference on Machine Learning},
  year={2024}
}

@inproceedings{gao2023pal,
  title={{PAL}: Program-aided Language Models},
  author={Gao, Luyu and Madaan, Aman and Zhou, Shuyan and Alon, Uri and Liu, Pengfei and Yang, Yiming and Callan, Jamie and Neubig, Graham},
  booktitle={International Conference on Machine Learning},
  year={2023}
}

@article{wu2023autogen,
  title={{AutoGen}: Enabling Next-Gen {LLM} Applications via Multi-Agent Conversation},
  author={Wu, Qingyun and Bansal, Gagan and Zhang, Jieyu and Wu, Yiran and Li, Beibin and Zhu, Erkang and Jiang, Li and Zhang, Xiaoyun and Zhang, Shaokun and Liu, Jiale and others},
  journal={arXiv preprint arXiv:2308.08155},
  year={2023}
}

@inproceedings{hong2024metagpt,
  title={{MetaGPT}: Meta Programming for A Multi-Agent Collaborative Framework},
  author={Hong, Sirui and Zhuge, Mingchen and Chen, Jonathan and Zheng, Xiawu and Cheng, Yuheng and Zhang, Ceyao and Wang, Jinlin and Wang, Zili and Yau, Steven Ka Shing and Lin, Zijuan and others},
  booktitle={International Conference on Learning Representations},
  year={2024}
}

@inproceedings{lewis2020retrieval,
  title={Retrieval-Augmented Generation for Knowledge-Intensive {NLP} Tasks},
  author={Lewis, Patrick and Perez, Ethan and Piktus, Aleksandra and Petroni, Fabio and Karpukhin, Vladimir and Goyal, Naman and K{\"u}ttler, Heinrich and Lewis, Mike and Yih, Wen-tau and Rockt{\"a}schel, Tim and others},
  booktitle={Advances in Neural Information Processing Systems},
  volume={33},
  year={2020}
}

@inproceedings{shinn2024reflexion,
  title={Reflexion: Language Agents with Verbal Reinforcement Learning},
  author={Shinn, Noah and Cassano, Federico and Gopinath, Ashwin and Narasimhan, Karthik and Yao, Shunyu},
  booktitle={Advances in Neural Information Processing Systems},
  volume={36},
  year={2024}
}

@inproceedings{madaan2024selfrefine,
  title={Self-Refine: Iterative Refinement with Self-Feedback},
  author={Madaan, Aman and Tandon, Niket and Gupta, Prakhar and Hallinan, Skyler and Gao, Luyu and Wiegreffe, Sarah and Alon, Uri and Dziri, Nouha and Prabhumoye, Shrimai and Yang, Yiming and others},
  booktitle={Advances in Neural Information Processing Systems},
  volume={36},
  year={2024}
}

@article{chen2023fireact,
  title={{FireAct}: Toward Language Agent Fine-tuning},
  author={Chen, Baian and Shu, Chang and Shareghi, Ehsan and Collier, Nigel and Narasimhan, Karthik and Yao, Shunyu},
  journal={arXiv preprint arXiv:2310.05915},
  year={2023}
}

@article{zeng2024agenttuning,
  title={{AgentTuning}: Enabling Generalized Agent Abilities for {LLMs}},
  author={Zeng, Aohan and Liu, Mingdao and Lu, Rui and Wang, Bowen and Liu, Xiao and Dong, Yuxiao and Tang, Jie},
  journal={arXiv preprint arXiv:2310.12823},
  year={2024}
}

@article{liu2025polymer,
  title={Harnessing large language models for data-scarce learning of polymer properties},
  author={Liu, Ning and Jafarzadeh, Siavash and Lattimer, Brian Y and Ni, Shuna and Lua, Jim and Yu, Yue},
  journal={Nature Computational Science},
  volume={5},
  number={3},
  pages={245--254},
  year={2025}
}

@article{liu2026beyond,
  title={Beyond Pairs: Your Language Model is Secretly Optimizing a Preference Graph},
  author={Liu, Ning and Sun, Chuanneng and Klinkner, Kristina and Malmasi, Shervin},
  journal={arXiv preprint arXiv:2605.08037},
  year={2026}
}

@inproceedings{zheng2024judging,
  title={Judging {LLM}-as-a-Judge with {MT-Bench} and Chatbot Arena},
  author={Zheng, Lianmin and Chiang, Wei-Lin and Sheng, Ying and Zhuang, Siyuan and Wu, Zhanghao and Zhuang, Yonghao and Lin, Zi and Li, Zhuohan and Li, Dacheng and Xing, Eric and others},
  booktitle={Advances in Neural Information Processing Systems},
  volume={36},
  year={2024}
}

@inproceedings{hu2022lora,
  title={{LoRA}: Low-Rank Adaptation of Large Language Models},
  author={Hu, Edward J and Shen, Yelong and Wallis, Phillip and Allen-Zhu, Zeyuan and Li, Yuanzhi and Wang, Shean and Wang, Lu and Chen, Weizhu},
  booktitle={International Conference on Learning Representations},
  year={2022}
}

@inproceedings{xiao2024flowbench,
  title={Flowbench: Revisiting and benchmarking workflow-guided planning for llm-based agents},
  author={Xiao, Ruixuan and Ma, Wentao and Wang, Ke and Wu, Yuchuan and Zhao, Junbo and Wang, Haobo and Huang, Fei and Li, Yongbin},
  booktitle={Findings of the Association for Computational Linguistics: EMNLP 2024},
  pages={10883--10900},
  year={2024}
}

@inproceedings{pei2025flow,
  title={Flow-of-action: Sop enhanced llm-based multi-agent system for root cause analysis},
  author={Pei, Changhua and Wang, Zexin and Liu, Fengrui and Li, Zeyan and Liu, Yang and He, Xiao and Kang, Rong and Zhang, Tieying and Chen, Jianjun and Li, Jianhui and others},
  booktitle={Companion Proceedings of the ACM on Web Conference 2025},
  pages={422--431},
  year={2025}
}

@article{nandi2025sop,
  title={Sop-bench: Complex industrial sops for evaluating llm agents},
  author={Nandi, Subhrangshu and Datta, Arghya and Vichare, Nikhil and Bhattacharya, Indranil and Raja, Huzefa and Xu, Jing and Ray, Shayan and Carenini, Giuseppe and Srivastava, Abhi and Chan, Aaron and others},
  journal={arXiv preprint arXiv:2506.08119},
  year={2025}
}

@article{wang2025sop,
  title={SOP-Maze: Evaluating Large Language Models on Complicated Business Standard Operating Procedures},
  author={Wang, Jiaming and Tang, Zhe and Jin, Zehao and Chen, Hefei and Jin, Yilin and Ding, Peng and Li, Xiaoyu and Cao, Xuezhi},
  journal={arXiv preprint arXiv:2510.08942},
  year={2025}
}

@inproceedings{kwon2023efficient,
  title={Efficient Memory Management for Large Language Model Serving with PagedAttention},
  author={Woosuk Kwon and Zhuohan Li and Siyuan Zhuang and Ying Sheng and Lianmin Zheng and Cody Hao Yu and Joseph E. Gonzalez and Hao Zhang and Ion Stoica},
  booktitle={Proceedings of the ACM SIGOPS 29th Symposium on Operating Systems Principles},
  year={2023}
}

@misc{zhao2024swiftascalablelightweightinfrastructure,
      title={SWIFT:A Scalable lightWeight Infrastructure for Fine-Tuning},
      author={Yuze Zhao and Jintao Huang and Jinghan Hu and Xingjun Wang and Yunlin Mao and Daoze Zhang and Zeyinzi Jiang and Zhikai Wu and Baole Ai and Ang Wang and Wenmeng Zhou and Yingda Chen},
      year={2024},
      eprint={2408.05517},
      archivePrefix={arXiv},
      primaryClass={cs.CL},
      url={https://arxiv.org/abs/2408.05517},
}

@misc{strands2025,
  title={Strands Agents SDK},
  author={{Strands Agents}},
  year={2025},
  howpublished={\url{https://github.com/strands-agents/harness-sdk}},
  note={Accessed: 2026-06-04}
}

@article{yuan2023scaling,
  title={Scaling relationship on learning mathematical reasoning with large language models},
  author={Yuan, Zheng and Yuan, Hongyi and Li, Chengpeng and Dong, Guanting and Lu, Keming and Tan, Chuanqi and Zhou, Chang and Zhou, Jingren},
  journal={arXiv preprint arXiv:2308.01825},
  year={2023}
}

@manual{aws_bedrock_agentcore_2026,
  title        = {Amazon Bedrock AgentCore Developer Guide},
  author       = {{Amazon Web Services}},
  year         = {2026},
  url          = {https://docs.aws.amazon.com/bedrock-agentcore/latest/devguide/what-is-bedrock-agentcore.html},
  note         = {Accessed: 2026-06-04}
}

@article{liu2024lost,
  title={Lost in the Middle: How Language Models Use Long Contexts},
  author={Liu, Nelson F. and Lin, Kevin and Hewitt, John and Paranjape, Ashwin and Bevilacqua, Michele and Petroni, Fabio and Liang, Percy},
  journal={Transactions of the Association for Computational Linguistics (TACL)},
  volume={12},
  pages={157--173},
  year={2024}
}

@inproceedings{kujanpaa2026toolmaking,
  author       = {Kalle Kujanp\"a\"a and Ning Liu and Shahnawaz Alam and Yeshwanth Reddy Sura and Tianyu Yang and Kristina Klinkner and Shervin Malmasi},
  title        = {Tool-Making and Self-Evolving {LLM} Agents in Low-Latency Systems},
  year         = {2026},
    booktitle = "Proceedings of the 2026 Conference on Empirical Methods in Natural Language Processing: Industry Track",
    address = "Budapest (Hungary)",
    publisher = "Association for Computational Linguistics",
}

@article{zheng2024sglang,
      title={SGLang: Efficient Execution of Structured Language Model Programs}, 
      author={Lianmin Zheng and Liangsheng Yin and Zhiqiang Xie and Chuyue Sun and Jeff Huang and Cody Hao Yu and Shiyi Cao and Christos Kozyrakis and Ion Stoica and Joseph E. Gonzalez and Clark Barrett and Ying Sheng},
  journal={Advances in neural information processing systems},
  volume={37},
  pages={62557--62583},
  year={2024}
}

\appendix
\section*{\centering Appendix}

\section{Tool Ecosystem Details}
\label{app:tools}

\paragraph{Code Interpreter.} The agent performs data manipulation and metric computation through a persistent Python interpreter (CodeAct~\cite{wang2024executable}) whose state (variables, imports, and intermediate results) is retained across tool invocations within an investigation. Operational data retrieved through MCP tools is analyzed programmatically to perform operations such as filtering, aggregating, and computing derived quantities, so that only the resulting view enters the model's input context rather than the full table. Executing comparisons and threshold checks in code eliminates arithmetic errors that arise when models perform these operations in natural language, and loops over CodeAct functions (programmatic wrappers exposed within the interpreter) let the agent batch repeated queries without issuing a separate tool call per iteration.

\paragraph{Task Tracking.} A \texttt{TodoList} tool, exposed within the interpreter, lets the agent record and update the steps of an investigation as it traverses the decision graph. Beyond bookkeeping, this serves as a steering mechanism: maintaining an explicit plan keeps the agent on the intended path through parallel branches, surfaces which nodes remain to be evaluated, and reduces premature termination on multi-step procedures.

\paragraph{MCP Tools.} Data retrieval and actions are implemented through the Model Context Protocol~\cite{anthropic2024mcp}, providing a standardized interface to warehouse operational dashboards. A \texttt{query\_metric} interface abstracts tool invocation, accepting a tool name, warehouse identifier, and time range, and returning results as pandas DataFrames for the interpreter to consume. Decoupling the agent's reasoning from data-source implementation lets dashboard backends evolve independently, and thread-safe connection management enables concurrent queries from parallel sub-agents.

\paragraph{Agentic RAG.} The agent augments its decision-making with domain knowledge retrieved on demand. Unlike static pipelines that prepend fixed context~\cite{lewis2020retrieval}, the agentic RAG component operates via tool-calling~\cite{schick2024toolformer}: when the agent detects a knowledge gap, it formulates a query and invokes a knowledge-base tool, receiving relevant passages as a tool response. This grounds reasoning in operational definitions, site-specific configurations, and historical incident documentation, and lets these definitions be updated without modifying the core SOP structure.

\paragraph{Memory.} The framework includes an operative memory that persists across agent invocations, accumulating knowledge over time rather than treating each investigation in isolation. The agent queries memory for prior findings on an entity before investigating and writes its outcome afterward, so observations build into patterns that are not visible within a single run. Deduplication follows as a direct consequence: when a prior invocation already resolved the same condition or determined that no remediation is possible, the current agent omits the redundant investigation. Memory entries expire after a configurable window so that stale findings do not suppress legitimate re-evaluation.

\section{Training Pipeline Details}
\label{app:training_pipeline}

\paragraph{Episodic Learning for Teacher Improvement.}
Initial teacher trajectories contain errors such as missed branches and threshold misinterpretations. Rather than manually correcting these, we iteratively improve the teacher through episodic learning~\cite{shinn2024reflexion} without weight updates. The episodic memory is used only during teacher trajectory generation and is excluded from student training and inference. After each batch of teacher runs, trajectories are evaluated against ground truth. For failures, a separate LLM analyzes the error and generates candidate memory entries that would prevent recurrence. New entries are added, and a consolidation step merges redundant entries to keep the memory compact. The teacher is re-run with updated memory injected into its prompt, iterating for $N$ rounds until convergence. Providing the memory at inference would yield equivalent accuracy (our ablation in Section~\ref{sec:ablation} confirms this), but adds prompt overhead and requires maintaining a growing memory store. Training on improved trajectories without the memory forces the student to internalize the episodic corrections in its weights, eliminating runtime context dependence.

\paragraph{Trajectory Decomposition and Fine-Tuning.}
Training trajectories are generated by running the teacher on labeled scenarios within the full framework (skills, MCP tools, code execution). For each scenario, the teacher queries operational data through the same tools used at inference time, producing a multi-turn trajectory of reasoning, tool calls, and tool outputs. Scenarios are sampled with balanced representation across outcome categories, and temporal partitioning enforces train-test separation. We decompose each trajectory into $T$ training samples (one per agent turn): the $t$-th sample takes as input the conversation history up to turn $t$ (system prompt, all prior tool calls and outputs, with the model's own reasoning from prior turns stripped) and targets the complete reasoning and action for turn~$t$. Retaining tool outputs is necessary: omitting them would desynchronize the model's state from the actual execution trace. Stripping prior reasoning forces independent generation at each turn, focusing the loss on single-turn decision quality. Not all trajectories are correct even after episodic improvement. We apply rejection sampling~\cite{yuan2023scaling} with LLM-judge evaluation~\cite{zheng2024judging}: numerical outputs are compared with 2\% tolerance, and categorical outputs require exact match. Only fully correct trajectories are retained for fine-tuning; exploiting the rejected trajectories through preference-based objectives~\cite{liu2026beyond} is a natural extension.

\section{Episodic Learning Ablation Details}
\label{app:ablation}

To isolate the impact of episodic learning and understand whether asymmetric distillation is necessary, an ablation is performed on Tasks~5 and~10 from the benchmark (Table~\ref{tab:ablation_el}). Four variants are tested: (1) no EL anywhere, (2) EL in teacher only (our approach), (3) EL memory provided to the student during both training and inference, and (4) EL memory provided during student training only, removed at inference. Applying EL only to the teacher (our approach) yields +16.2\% on Task~5 and +28.6\% on Task~10 over the no-EL variant. When EL is given to the student at both training and evaluation, comparable accuracy is reached (+16.6\%, +25.3\%), but an inference-time dependence on the growing EL context is introduced, adding latency for no accuracy gain. Training with EL but evaluating without it degrades catastrophically ($-$52.7\%, $-$40.0\%), falling below even the no-EL variant. This shows that naively exposing the student to episodic context creates a brittle dependency. Our asymmetric design distills the episodic corrections into the student's weights, removing any inference-time memory requirement.

\section{Cost Analysis}
\label{app:cost}

Training and deploying a capable model is an unavoidable investment when the goal is faithful execution of complex, multi-step SOPs with strict correctness requirements: off-the-shelf models plateau regardless of scale (OTS Qwen3-235B averages $-$10\% versus baseline, and even the 355B teacher reaches only +6.1\%; Table~\ref{tab:benchmark_main}), consistent with the broader observation that general-purpose LLMs require domain-specific adaptation to perform reliably in specialized domains~\cite{liu2025polymer}, while the alternative of manual execution at ${\approx}6$ minutes per ticket carries far higher ongoing operational cost. We nonetheless take deliberate steps to minimize compute requirements. First, LoRA fine-tuning~\cite{hu2022lora} rather than full-parameter training reduces trainable parameters by orders of magnitude, completes in at most 2 epochs, and its rank can be adjusted to available resources; even a rank-32 fine-tune (\agentname{}-G) outperforms all OTS baselines. Second, asymmetric distillation compresses the 355B teacher into production-viable students (32B/106B) that match or exceed teacher accuracy, and the teacher is used only offline for trajectory generation, never at serving time. Third, at inference the fine-tuned 32B model is served on 8 H100 GPUs via vLLM with a 54\% latency reduction versus the teacher at equivalent accuracy, and merged LoRA weights add zero inference overhead. At production scale these recurring savings, e.g. ${\approx}8{,}000$ autonomously processed conveyance triggers with a 300\% resolution speedup or ticket processing below the manual rate (median 187s vs.\ ${\approx}360$s), substantially exceed the one-time training cost.

\section{Training and Inference Details}
\label{app:training_details}

Models are fine-tuned using MS-SWIFT~\cite{zhao2024swiftascalablelightweightinfrastructure} with LoRA~\cite{hu2022lora}. For Qwen3-32B, we use rank and alpha of 256, learning rate $2{\times}10^{-4}$, and train on 64 H100 GPUs. For GLM-4.5-Air, we use rank and alpha of 32, learning rate $1{\times}10^{-4}$, and train on 128 H100 GPUs. Both use a batch size of 64, 5\% linear warmup followed by cosine annealing to zero, and train for 2 epochs. The ticket-processing model \agentname{}-G-XL fine-tunes GLM-4.7 (355B) with the same MS-SWIFT/LoRA pipeline (rank and alpha 64, learning rate $5{\times}10^{-5}$, batch size 64, 1 epoch) on 128 H100 GPUs. LoRA weights are merged after training, eliminating inference overhead. At inference, \agentname{}-G and \agentname{}-Q are served via vLLM~\cite{kwon2023efficient} on 8 H100 GPUs, and \agentname{}-G-XL via SGLang~\cite{zheng2024sglang} on 8 H200 GPUs.

\section{Ticket Processing Evaluation}
\label{app:ciss}

\paragraph{Model.} The deployed ticket-processing agent is \agentname{}-G-XL, an open-weight GLM-4.7 (355B) fine-tuned with the same trajectory-centric pipeline as \agentname{}-G and \agentname{}-Q (Appendix~\ref{app:training_pipeline}, config in Appendix~\ref{app:training_details}). Ticket processing runs on a longer interval than robotic conveyance, so a larger fine-tune is admissible: \agentname{}-G-XL's median generation duration (Section~\ref{subsec:ticket}) meets that requirement. All results in Table~\ref{tab:ciss_results} are from this single fine-tuned model.

\paragraph{Evaluation setup.} The 410 tickets are an unbiased sample of open consolidation tickets at one fulfillment center on a single day, processed by the agent in the production framework with live inventory access. The task has no deterministic ground truth (the correct solution depends on live inventory state at processing time), so correctness is estimated by expert review of agent traces: three experienced operations associates per batch examined each original ticket and its correspondence alongside the agent's reasoning trace, and judged the solution as correct or incorrect. Tickets were distributed across reviewers to maximize coverage rather than labeled redundantly, so by design no inter-rater agreement statistic exists for this review.

\end{document}